\def\year{2022}\relax
\documentclass[letterpaper]{article} 
\usepackage{aaai22}  
\usepackage{times}  
\usepackage{helvet}  
\usepackage{courier}  
\usepackage[hyphens]{url}  
\usepackage{graphicx} 
\usepackage{bm}
\usepackage{booktabs}
\usepackage{amsmath}
\usepackage{bbm}
\usepackage{url}
\usepackage{newfloat}
\usepackage{algorithm}
\usepackage{algorithmic}
\usepackage{amssymb}
\usepackage{listings}
\usepackage{bm}
\usepackage{natbib}  
\usepackage{caption} 
\DeclareCaptionStyle{ruled}{labelfont=normalfont,labelsep=colon,strut=off} 
\usepackage{algorithm}
\usepackage{algorithmic}

\usepackage{newfloat}
\usepackage{listings}
\floatstyle{ruled}
\newfloat{listing}{tb}{lst}{}
\floatname{listing}{Listing}
\title{Imagine by Reasoning: A Reasoning-Based Implicit Semantic Data Augmentation for Long-Tailed Classification}
\author{
	Xiaohua Chen\textsuperscript{\rm 1,2}, 
	Yucan Zhou\textsuperscript{\rm 1}\thanks{Yucan Zhou is the corresponding author}, 
	Dayan Wu\textsuperscript{\rm 1}, \\
	Wanqian Zhang\textsuperscript{\rm 1}, 
	Yu Zhou\textsuperscript{\rm 1}, 
	Bo Li\textsuperscript{\rm 1,2}, 
	Weiping Wang\textsuperscript{\rm 1,2}
}
\affiliations{
	\textsuperscript{\rm 1} Institute of Information Engineering, Chinese Academy of Sciences, Beijing, China\\
	\textsuperscript{\rm 2} School of Cyber Security, University of Chinese Academy of Sciences, Beijing, China\\
	\{chenxiaohua, zhouyucan, wudayan, zhangwanqian, zhouyu, libo, wangweiping\}@iie.ac.cn
	
}

\usepackage{bibentry}

\begin{document}
	\maketitle
	\begin{abstract}
		Real-world data often follows a long-tailed distribution, which makes the performance of existing classification algorithms degrade heavily. A key issue is that samples in tail categories fail to depict their intra-class diversity. Humans can imagine a sample in new poses, scenes, and view angles with their prior knowledge even if it is the first time to see this category. Inspired by this, we propose a novel reasoning-based implicit semantic data augmentation method to borrow transformation directions from other classes. Since the covariance matrix of each category represents the feature transformation directions, we can sample new directions from similar categories to generate definitely different instances. Specifically, the long-tailed distributed data is first adopted to train a backbone and a classifier. Then, a covariance matrix for each category is estimated, and a knowledge graph is constructed to store the relations of any two categories. Finally, tail samples are adaptively enhanced via propagating information from all the similar categories in the knowledge graph. Experimental results on CIFAR-100-LT, ImageNet-LT, and iNaturalist 2018 have demonstrated the effectiveness of our proposed method\footnote{Code is available at \url{https://github.com/xiaohua-chen/RISDA}} compared with the state-of-the-art methods. 
	\end{abstract}
	
	\section{Introduction}
	Deep neural networks have achieved dramatic performance when high-quality annotated and balanced distributed training data is provided \cite{KrizhevskySH12, he2016deep, HuangLMW17}. However, in real-world scenarios, data is usually long-tail distributed, where many tail categories occupy only a small number of samples and most samples belong to a few head categories. It is a great challenge to train deep models on long-tail distributed data. On the one hand, the separating hyperplane will be heavily skewed to the tail classes because of their weak statistical ability. More importantly, tail classes are easily overfitted as their samples fail to describe their intra-class diversity.

	
	
	To strengthen the effect of tail classes in training, a typical and intuitive strategy is re-balancing, including data re-sampling and loss re-weighting. Data re-sampling
	achieves training fairness by balancing the data distribution through sampling, while loss re-weighting pays more attention to tail categories by assigning higher penalties to the loss of tail classes. Although these methods can alleviate hyperplane skewness, over-fitting on tail categories is still a big issue because of the limited intra-class diversity. 
	
	\begin{figure}
		\setlength{\abovecaptionskip}{0pt}
		\includegraphics[width=\linewidth,scale=1.00]{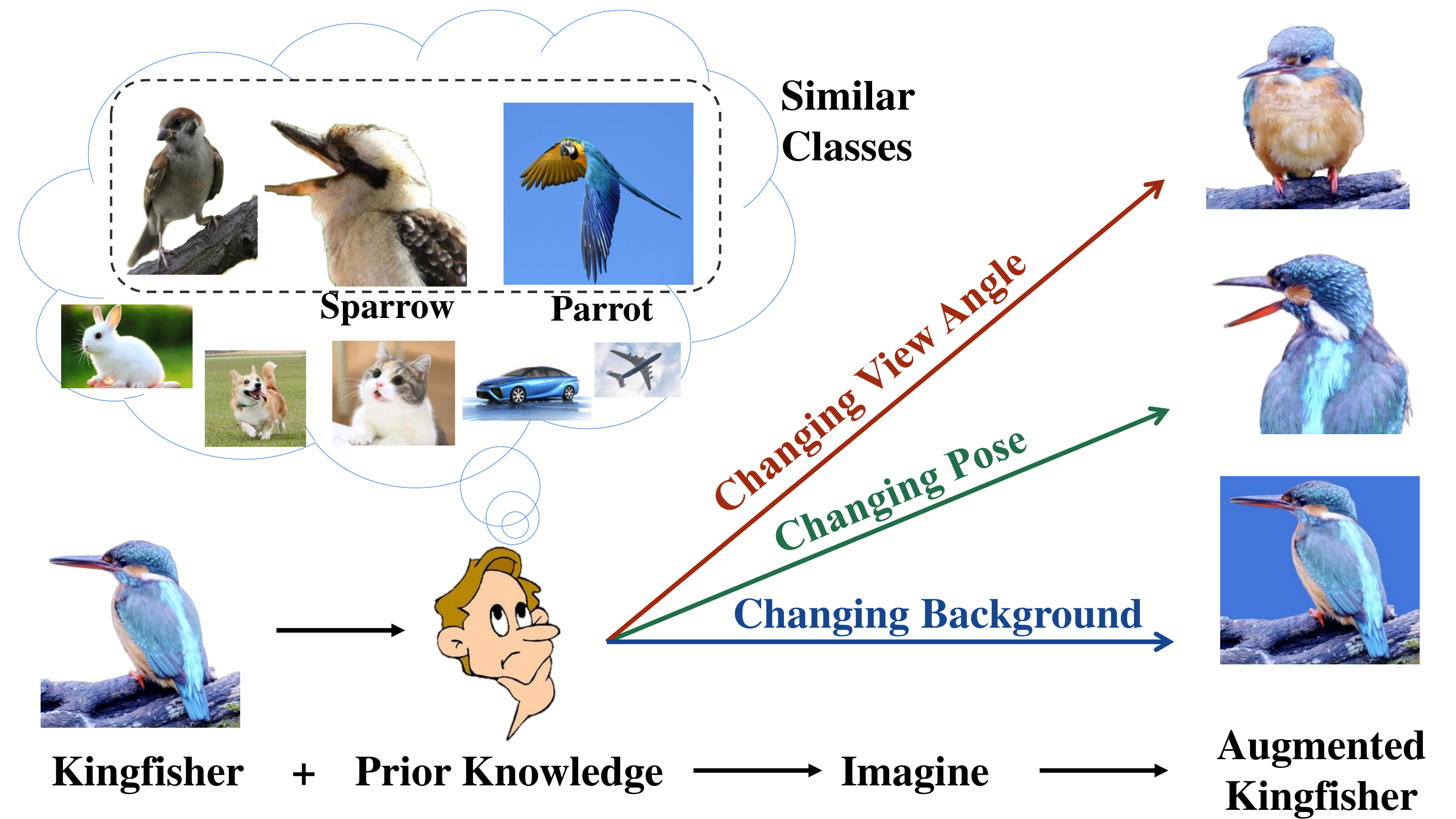}
		\caption{An illustration of imagining new samples by reasoning.
			People can imagine a kingfisher in various view angles, poses, and backgrounds with their prior knowledge.  }
		\label{fig:EP}
	\end{figure}
	
	Data augmentation is an efficient way to enrich tail categories, where cropping, rotation, mixup, and generative adversarial network are adopted to generate new samples \cite{simonyan2014very,he2016deep,ratner2017learning,bowles2018gan}. Nevertheless, with the increasing number of augmented examples, the training speed is sharply slowed down. 
	Fortunately, ISDA \cite{wang2019implicit} propose an instance-based implicit data augmentation method to reduce the computation cost, where new instances can be generated by changing the original instance to semantic transformation directions sampled from the feature covariance matrix. However, for tail classes, it is incapable to estimate a diversified covariance matrix
	\cite{li2021metasaug}. 
	
	We argue that humans can imagine a rare animal in different poses, colors, and backgrounds with prior knowledge, though we have only seen one picture of this animal. For instance (shown in Figure \ref{fig:EP}), as ``Kingfisher" is similar to ``Sparrow" in humans' prior knowledge, we can easily imagine a ``Kingfisher" with an open mouth when we have seen a ``Sparrow" with an open mouth. Inspired by this, tail classes can borrow semantic transformations from other classes.

	In this paper, in order to mimic the human reasoning process, we propose an effective Reasoning-Based Implicit Semantic Data Augmentation (\textbf{RISDA}) method for long-tailed classification. It enriches a tail instance by transferring semantic transformation directions from similar categories to expand the intra-class diversity. Specifically, we first train a network with the long-tail distributed data. Then, a covariance matrix for each category is estimated, which represents all the possible transformation directions of a category. After that, for each tail category whose covariance matrix is limited, we enhance it with similar categories. To recognize similar categories, a knowledge graph is constructed, where each non-diagonal element indicates the similarity of two categories. Finally, sufficient instances can be generated by augmenting a tail instance via propagating transformation directions from similar categories. Nevertheless, to ensure the tail instance contains all the features to transfer, we also complement it with features from similar categories. Therefore, our method can generate instances that are definitely different from the training data. 
	The contributions of this paper are summarized as follows.
	\begin{itemize}
		\item We propose a reasoning-based implicit data augmentation method, which transfers semantic transformation directions from other classes to enhance tail classes. It can largely enrich the intra-class diversity for tail categories. To the best of our knowledge, this is the first time to transfer feature transformations from similar categories for augmentation. 
		\item We construct a learnable knowledge graph so that our method can adaptively select similar categories for different samples to make the reasonable transformation.
		\item The proposed RISDA outperforms current state-of-the-art methods on long-tailed datasets, such as CIFAR-100-LT, ImageNet-LT, and iNaturalist 2018.
	\end{itemize}

	\section{Related Work}
	\subsection{Re-Balance}
	A common re-balance strategy is re-sampling, which aims to achieve a balanced data distribution, including over-sampling \cite{buda2018systematic,byrd2019effect} and under-sampling \cite{he2009learning,buda2018systematic}. However, over-sampling may cause over-fitting by duplicating tail data, while under-sampling will damage feature representation when abandoning head data. 
	Recently, some decoupling methods are developed, which adopt different sampling strategies in representation learning and classifier training\cite{kang2019decoupling,zhou2020bbn,wang2021contrastive}.
	
	Another strategy is re-weighting the loss to give more attention to tail classes. An intuitive way is to re-weight the loss of each class inversely proportional to the number of samples \cite{huang2016learning}. Then, class-balance loss \cite{cui2019class} is proposed to emphasize the effective number of samples. Thereafter, meta-class-weight \cite{jamal2020rethinking} estimates class weights by meta-learning, while LDAM \cite{cao2019learning} uses class-level re-weighting optimization schedule to train a label-distribution-aware loss.
	In addition, some fine-grained instance-level re-weighting methods are studied. Such as Focal Loss \cite{lin2017focal} reduces the weights of easy samples to make the model focus on hard samples in training. L2RW \cite{ren2018learning} and meta-weight-net \cite{shu2019meta} assign instance-wise weights on the gradient direction. 
	
	Although re-balance methods have achieved great improvements, the intra-class diversity for tail categories is still limited, making them easily over-fitted.
	
	\subsection{Data Augmentation}
	To enhance the intra-class diversity for tail categories, data augmentation methods are explored. A direct way is to generate samples by combining other samples in a tail class linearly \cite{he2008adaptive,zhang2017mixup}. However, the combination may be meaningless. To address this issue, GAN \cite{bowles2018gan} is introduced. The above methods are explicit data augmentation methods. With the increasing number of samples, the convergence speed is sharply slowed down. ISDA \cite{wang2019implicit} is a great way to handle this problem, which uses the class-wise covariance matrix and instance-wise feature to formulate a Gaussian distribution, thus infinite samples can be generated. But ISDA fails for tail classes where the insufficient samples are incapable of estimating a diversified covariance matrix. MetaAug \cite{li2021metasaug} tries to find an optimal covariance matrix that generalizes well on unseen samples. However, it still suffers from limited diversity for tail classes as the covariance matrix is calculated with samples at hand.
	
	Fortunately, tail classes can borrow information from other categories \cite{chu2020feature,liu2020deep,xiao2021does}. As head categories are sufficiently diversified, some methods propose to transfer knowledge from head to tail. For example, Liu et al. model each category into a ``feature cloud” and expand tail categories by transferring the intra-class angular distribution from head categories \cite{liu2020deep}. Chu et al. generate new high-level features for tail categories by merging class-specific features of the head categories and class-generic features of the tail categories \cite{chu2020feature}. However, the ``head to tail" is defined by the number of samples, transferring variations based on this relation may be meaningless. For example, if ``Apple"  is a head class, and ``Kingfisher'' is a tail class, then transferring color variations from ``Apple" to ``Kingfisher'' is unreasonable.
	
	Our method follows the idea of implicit data augmentation and tries to borrow variations from other categories to enrich tail categories. However, different from other methods, we transfer variations from all the related classes in a knowledge graph other than the head categories. 
	
	
	\section{The Proposed Method}
	\begin{figure*}[!tp]
		\centering
		\includegraphics[width=\linewidth,scale=1.00]{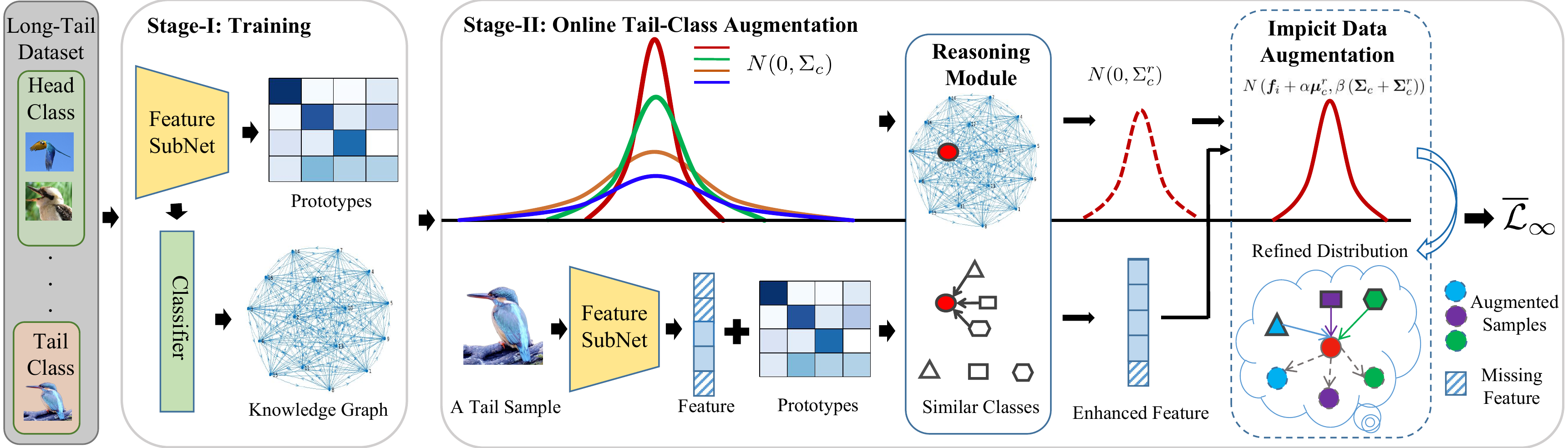}
		\caption{The framework of RISDA. In Stage-I, all training data is used to train a feature subnet and a classifier. Subsequently, we calculate the covariance matrix and the prototype for each category with the feature extracted with the subnet. Then, the classifier is used to construct a knowledge graph. In Stage-II, we augment each tail instance with a semantic transformation direction sampled from the covariance matrix. For tail classes, the covariance matrix is limited, so we refine it with similar classes defined in the knowledge graph. What's more, to ensure the tail instance contains all the features to transfer, we complement it with the prototypes from similar categories. Finally, unlimited instances can be sampled to calculate the $\mathcal{\overline{L}_{\infty}}. $ 
		}
		\label{Architecture}
	\end{figure*}
	A classifier usually performs worse for tail classes. To enhance their diversity, we propose to borrow variations from other categories as shown in Figure \ref{Architecture}.
	
	In stage-I, we use all the samples to train a feature subnetwork and a classifier. Subsequently, we use the extracted features to calculate the covariance matrix and the prototype for each category, where the covariance matrix represents all the feature semantic transformation directions of each category, and the prototype contains its specific features. Then, the classifier is adopted to construct a knowledge graph, which stores the similarity of any two categories. In stage-II, a tail sample is first propagated to the feature subnetwork to get semantic features. Then, we aim to augment it with the semantic transformations sampled from the covariance matrix of its category. However, the covariance matrix of tail categories is limited, so we refine it with similar classes defined in the knowledge graph. What's more, the tail instance may miss some features to transfer. Thus, to ensure it contains all the features, we complement it with the features from similar categories. Consequently, unlimited instances can be generated by changing an instance with infinite semantic transformation directions from the refined distribution. Finally, an upper bound of the CE loss on unlimited data is derived to fine-tune the network.
	
	In the following, we will describe how a tail sample is implicitly augmented by reasoning, and the loss function to fine-tune the network with unlimited generated data.
	
	\subsection{Implicit Data Augmentation}
	Given a long-tail distributed dataset $\left\lbrace x_i,y_i\right\rbrace_{i=1}^N$ , where $y_i \in L=\left\lbrace {l_{1},l_{2},...,l_{C} }\right\rbrace$, $C$ is the number of categories. Inspired by the observation that instance-balanced sampling learns better generalizable representations \cite{kang2019decoupling}, we use the original distribution to train a feature extractor $F$ and a classifier $H$ in Stage-I. 
	
	For a sample $x_i$ in a tail class $l_c$, we extract its feature with $\bm{f}_i=F(x_i)$. Inspired by ISDA, we can randomly sample along $N \left(0 ,\beta \bm{  \Sigma}_c \right)$ to generate features with different semantic transformations,  i.e., $\tilde{\bm{f}}_{i}    \sim N \left(\bm{f}_{i} ,\beta \bm{  \Sigma}_c \right)$.
	%
	
	As the semantic transformation directions of different categories are different, we need to compute a covariance matrix for each category. To get the covariance matrix, we first need to calculate each class prototype by averaging features of all the samples in a class, as shown in Equation (\ref{eq1}): 
	
	\begin{equation}
	\label{eq1}
	\bm{\mu}_{c}=\dfrac{1}{N_{c}} \sum_{i=1}^{N_{c}}\bm{f}_{i},
	\end{equation}
	where $c\in[1,C]$, $N_{c}$ is the number of samples in class $l_c$, and $\bm{f}_{i}$ represents feature of the $i$-th sample in $l_c$. Secondly, we calculate each element in a covariance matrix with: 
	
	\begin{equation}
	\label{eq2}
	\bm{\Sigma}_{c}\left(m,n\right)= \frac{ \bm{\Sigma}^{N_{c}}_{i=1}\left(\bm{f}_{i}^{m}-\bm{\mu}_c^{m}\right) \left(\bm{f}_{i}^{n}-\bm{\mu}_c^{n}\right) }{N_c-1}.
	\end{equation}
	After that, the covariance matrixes of all classes $\bm{\Sigma}=\left\lbrace \bm{\Sigma}_{1}, \bm{\Sigma}_{2},\cdots,\bm{\Sigma}_{C}  \right\rbrace $ can be obtained. 
	However, covariance matrixes for tail classes contain limited variations.
	To alleviate this issue, we propose a novel reasoning module to transfer plentiful variations from other classes.

	\subsection{Reasoning-Based Implicit Data Augmentation}
	In this section, we elaborate on the process of transferring knowledge to augment a tail sample. With the constructed knowledge graph, we can enrich the covariance matrixes of tail classes by reasoning.
	\subsubsection{Knowledge Graph Construction}
	Current methods usually transfer variations from head classes to the tail ones. Instead, our approach tries to transfer knowledge from similar categories. To achieve this, we use the classifier $H$ to construct a knowledge graph, which is built with the confusion matrix on the training data. Actually, it is a category-to-category directed graph $ \left\langle  V, \bm{\varepsilon}\right\rangle $, where $V$ are category nodes and $\bm{\varepsilon} \in R^{C\times C}$. Each element $\varepsilon_{ij}$ represents the similarity of $l_i$ and $l_j$, which can be calculated as follows:
	
	\begin{equation}
	\label{eq4}
	\varepsilon_{ij}=\frac{\sum_{i=1}^{N_i}{\mathbbm{1}\left(H(\bm{f}_i)=l_j \land y_i=l_i\right)}}{N_i},
	\end{equation}
	where $\mathbbm{1}$ is the indicator function. $\varepsilon_{ij}$ represents the ratio of the number of samples in $l_i$ predicted into $l_j$.
	
	\subsubsection{Reasoning-Based Semantic Transformation}
	We propose to refine the covariance matrix of each tail class with those of its similar classes. Intuitively, different categories are similar because they have similar appearances or living environments. 
	%
	For example, because ``Kingfisher" and ``Parrot" are very similar, when we see a flying ``Parrot", we can imagine a flying ``Kingfisher" even if we haven't seen a flying ``Kingfisher".
	Therefore, we can infer a better intra-class covariance for tail classes by transferring semantic transformations from similar classes:  
	\begin{equation}
	\label{eq6}
	\bm{\Sigma}_{c}^{r}= \sum_{i=1,i \ne c}^{C} \varepsilon_{c,i} \bm{\Sigma}_{i}.
	\end{equation}
	Ideally, we need to compute the reasoning-based class-conditional covariance matrix with entire training samples in each epoch. It is cost-expensive. So we use an online process to update and transfer covariance matrix batch by batch.
	
	With the reasoning-based covariance matrix, we can change an instance with a more diversified semantic direction. However, an instance may miss some features that are necessary for a reasoning-based transformation direction. For example, only when the mouth is detected in a ``Kingfisher", we can change the pose of its mouth. Therefore, to ensure a tail instance contains all the features to transfer, we complement it with features from similar categories: 
	\begin{equation}
	\label{eq5}
	\bm{\mu}_{c}^{r}=\sum_{i=1,i\ne c}^{C} \varepsilon_{c,i} \bm{\mu}_{i}.
	\end{equation} 
	
	After the feature distribution of each tail sample is refined by a reasoning prototype and a reasoning covariance matrix.
	Then, $\bm{f}_i$ can perform various semantic transformations along the random direrctions sampled from $N \left(\alpha \bm{\mu}_{c}^r,\beta \left( \bm{  \Sigma}_{c}+\bm{  \Sigma}^{r}_{c} \right)\right)$ and generate the augmented feature $ \tilde{\bm{f}}_{i}  $ during trainning with Equation (\ref{eq7}): 
	\begin{equation}
	\label{eq7}
	\tilde{\bm{f}}_{i}    \sim N \left(\bm{f}_{i} +\alpha \bm{\mu}_{c}^r,\beta \left( \bm{  \Sigma}_{c}+\bm{  \Sigma}^{r}_{c} \right)\right),
	\end{equation}
	where $\alpha$ and $\beta$ are positive coefficients to control the strength of reasoning-based semantic data augmentation. In the experiment, both $\alpha$ and $\beta$ are decayed by $t/T$, where $t$ and $T$ represent the current number of epochs and the number of total epochs, respectively.
	\subsection{Optimization}
	With our data augmentation method, a straightforward way to train a classifier is to generate samples for tail classes with Equation (\ref{eq7}) untill they have comparable samples with head classes. Assume each sample in tail class $l_c$ is augmented $M$ times, then we can obtain a new data set $\left\lbrace \left(\bm{f}_{i}^1,y_{i} \right), \left(\bm{f}_{i}^2,y_{i} \right),...,\left(\bm{f}_{i}^M,y_{i} \right) \right\rbrace_{i=1}^{N_{c}}$, where $\bm{f}_{i}^{k}$ sampled from $\tilde{\bm{f}}_{i}$ is the $k$-th augmented feature of $\bm{f}_{i}$.
	
	Thereafter, we can use the traditional cross-entrory loss to train a classifier: 
	
	\begin{equation}
	\begin{split}
	\label{eq8}
	&\mathcal{L_{M}}(\bm{\theta}_{F},\bm{W},b)=\\
	&\sum_{c \in TC}\dfrac{1}{N_{c}}\sum_{i=1}^{N_c} \dfrac{1}{M}  
	\sum_{k=1}^{M}   -\log\left(  \dfrac{e^{\bm{w}_{y_{i}}^{T} \bm{f}_{i}^{k} +b_{y_{i}}   }    }{\sum_{j=1}^Ce^{\bm{w}_{j}^{T}\bm{f}_{i}^{k} +b_{j}} } \right) \\
	&+\sum_{c \in HC}\dfrac{1}{N_{c}}\sum_{i=1}^{N_c} 
	-\log\left(  \dfrac{e^{\bm{w}_{y_{i}}^{T}\bm{f}_{i} +b_{y_{i}}   }    }{\sum_{j=1}^Ce^{\bm{w}_{j}^{T}\bm{f}_{i} +b_{j}} }    \right),
	\end{split}
	\end{equation}
	where $TC$ represents the set of tail classes, $HC$ is the set of head classes,
	$\bm{W}=[\bm{w}_{1},\bm{w}_{2},...,\bm{w}_{C}]^{T}$ and $\bm{b}=[b_{1},b_{2},...,b_{C}]^{T} $ are the weight matrixes and
	biases corresponding to the last fully connected layer, respectively.
	\begin{table*}[!pt]
		\centering
		\begin{tabular}{lccccc}
			\toprule
			Imbalance factor ($\lambda$)      & 200   & 100   & 50    & 20    & 10 \\
			\midrule
			Cross-Entropy$^{\dagger}$      & 65.30 & 61.54 & 55.98 & 48.94 & 44.27 \\ 
			Mixup$^{\dagger}$ \cite{zhang2017mixup}     &-       &60.46       &55.01       &-       &41.98       \\ 
			L2RW$^{\dagger}$ \cite{ren2018learning}        &67.00       &61.10       &56.83       &49.25       &47.88       \\ 
			Class-balanced CE$^{\dagger}$ \cite{cui2019class}         &64.44       &61.23       &55.21       &48.06       &42.43       \\ 
			Class-balanced  fine-tuning$^{\dagger}$ \cite{DBLP:conf/cvpr/CuiSSHB18}         &61.34       &58.50       &53.78       &47.70       &42.43       \\ 
			Meta-weight net$^{\dagger}$ \cite{shu2019meta}                           &63.38      &58.39       &54.34      &46.96       &41.09       \\ 
			Meta-class-weight with cross-entropy loss$^{\dagger}$ \cite{jamal2020rethinking} &60.69       &56.65       &51.47       &44.38       &40.42       \\ 
			BBN$^{\ast}$ \cite{zhou2020bbn}                                       &-       &57.44       &52.98       &-       &40.88       \\ 
			Hybrid-PSC$^{\ast}$ \cite{wang2021contrastive}                                &-       &55.03       & 51.07      &-      &37.63       \\ 
			Bag of Tricks$^{\ast}$ \cite{zhang2021bag}                           &-       &52.17       &48.31       &-       &-       \\ 
			MetaSAug with cross-entropy loss$^{\ast}$ \cite{li2021metasaug}          &60.06       &53.13       &48.10       &42.15       &38.27       \\ 
			\textbf{RISDA}                                 &\textbf{55.24}       &\textbf{49.84}       &\textbf{46.16}       &\textbf{41.33}         &\textbf{37.62}       \\ 
			\bottomrule
		\end{tabular}
		\caption{Error rate ($\%$) of ResNet-32 on CIFAR-100-LT under different imbalance factors.}
		\label{tab1}
	\end{table*}

	However, for the tail category, when sampling $M$ times, the sampling variance is unstable and limited. An ideal way is to generate as much data as possible (i.e., set $M$ as large as possible). But the increasing $M$ can lead to additional calculations. To simplify computation while generating more data, we plan to implicitly generate unlimited features. When $M$ is close to infinity, we can derive an easy upper-bound loss following the Law of Large Numbers.
	In detail, when $M \rightarrow \infty$, we take into account all possible enhanced samples, and then the loss for tail categories can be defined as:
	\begin{equation}
	\label{eq9}
	\begin{aligned}	   
	&\mathcal{ L_{M \to \infty} } =\sum_{c \in {TC}}\dfrac{1}{N_{c}} \sum_{i=1}^{N_{c}} E_{  
		\tilde{\bm{f}}_{i}     } \left[-log\left(  \dfrac{e^{\bm{w}_{y_{i}}^{T}     \tilde{\bm{f}}_{i}+b_{y_{i}}   }  }{\sum_{j=1}^Ce^{\bm{w}_{j}^{T}\tilde{\bm{f}}_{i}   +b_{j}} }        \right)  \right] \\
	&=\sum_{c \in {TC}}\dfrac{1}{N_{c}} \sum_{i=1}^{N_{c}} E_{   \tilde{\bm{f}}_{i}   }\left[   log(    \sum _{j=1} ^{C} e^{(\bm{w}_{j}^{T}- \bm{w}^{T}_{y_{i}}    )  \tilde{\bm{f}}_{i}+(b_{j}-b_{y_{i}}
		)   } \right].
	\end{aligned}
	\end{equation}
	
	Nevertheless, the above equation is difficult to calculate accurately. With the help of Jensen’s inequality $E[logX] \leq logE[X]$, we can derive its upper bound as follows: 
	\begin{equation}
	\label{eq10}
	\begin{aligned}
	&\mathcal{L_{M \to \infty}} \leq \\
	&\sum_{c \in {TC}}\dfrac{1}{N_{c}} \sum_{i=1}^{N_{c}}
	log\left(    E_{    \tilde{\bm{f}}_{i}    }\left[     \sum _{j=1} ^{C} e^{(\bm{w}_{j}^{T}- \bm{w}^{T}_{y_{i}}    )   \tilde{\bm{f}}_{i}   +(b_{j}-b_{y_{i}}) }  \right] \right)  \\
	&=\sum_{c \in {TC}}\dfrac{1}{N_{c}} \sum_{i=1}^{N_{c}}
	log\left( \sum _{j=1} ^{C}   E_{    \tilde{\bm{f}}_{i}    }   \left[    e^{(\bm{w}_{j}^{T}- \bm{w}^{T}_{y_{i}}    )   \tilde{\bm{f}}_{i}    +(b_{j}-b_{y_{i}}) 
	}  \right] \right).
	\end{aligned}
	\end{equation}
	Because of 
	$  \tilde{\bm{f}}_{i}   \sim  N  \left( \bm{f}_{i} +\alpha \bm{\mu}_{y_i}^{r},\beta \left(  \bm{\Sigma}_{y_i}+\bm{\Sigma}^{r}_{y_i} \right) \right) $, we can obtain that $( \bm{w}_{j}^{T}-\bm{w}_{y_{i}}^{T}  ) \tilde{\bm{f}}_{i}  
	+(b_{j}-b_{y_{i}}) $ is also a Gaussian random variable, i.e.
	$( \bm{w}_{j}^{T}-\bm{w}_{y_{i}}^{T}  ) \tilde{\bm{f}}_{i}  
	+(b_{j}-b_{y_{i}})  \sim  N \left(    
	( \bm{w}_{j}^{T}-\bm{w}_{y_{i}}^{T}  )(   \bm{f}_{i} +\alpha \bm{\mu}_{y_i}^r)+(b_{j}-b_{y_{i}}) ,\sigma_{i}^{j}
	\right) $, where $\sigma_{i}^{j} =\beta(\bm{w}_{j}^{T}-\bm{w}_{y_{i}}^{T} ) \left( \bm{\Sigma}_{y_i}+\bm{\Sigma}_{y_i}^{r} \right)  (\bm{w}_{j}-\bm{w}_{y_i})$.
	
	After that, we use the moment-generating function: 
	
	\begin{equation}
	\label{eq11}
	\mathbb{E} [e^{tX}]=e^{t\mu +\dfrac{1}{2} \sigma t^{2} } , X \sim N(\mu,\sigma) 
	\end{equation} 
	Finally, we can get Equation (\ref{eq12}): 
	
	\begin{equation}
	\label{eq12}
	\mathcal{L_{M \to \infty}}  \leq \mathcal{\overline{L}_{\infty}     }=-\sum_{c \in {TC}} \dfrac{1}{N_{c}} \sum_{i=1}^{N_{c}}
	log\dfrac{e^{Z_{i}^{ y_{i}}} }  {\sum_{j=1}^{C} e^{Z_{i}^{j}}},
	\end{equation} 
	where 
	$Z_{i}^{j}=\hat{y}_{i}^{j}+\alpha(\bm{w}_{j}^{T} -\bm{w}_{y_{i}}^{T}    ) \bm{ \mu}_{j}^r + \dfrac{\sigma_{i}^{j}}{2}$, and $\hat{y}_{i}^{j}$ is the $j$-th logits of the output for $x_{i}$. Essentially, Equation (\ref{eq12}) provides a surrogate
	loss for our reasoning implicit data augmentation method, through optimizing upper bound $\mathcal{\overline{L} }$
	instead of minimizing the exact loss function $\mathcal{ L_{\infty}}$. This new loss can effectively adjust the classification decision boundary with the reasoning-based augmented features. 
	
	But when we generate unlimited samples for tail categories, the data distribution is still unbalanced where the head categories become ``tail categories". Therefore, similar to the tail class, we also make infinite enhancements for head classes. In this way, we find that head categories are still dominant because the augmentation results rely on the training samples and the head class has most of the samples. To solve this problem, we introduce the re-weighting strategy by setting different weights to different classes, which are defined as $\rho_{c} \approx (1-\gamma)/(1-\gamma^{N_{c}})$, where $\gamma$ is a hype-parameter. With the re-weighting strategy, we can modify the loss function as follows: 
	\begin{equation}
	\label{eq13}
	\mathcal{L_{M \to \infty}}   \leq \mathcal{\overline{L}_{\infty}     }=
	-\sum_{c=1}^{C} \dfrac{1}{N_{c}} \sum_{i=1}^{N_{c}}
	\rho_{c}log\dfrac{e^{Z_{i}^{ y_{i}}} }  {\sum_{j=1}^{C} e^{Z_{i}^{j}}}.
	\end{equation} 
	
	Our reasoning module can generate more diversified samples by reasoning features and transformations from other classes, which is shown in the experiment. In addition, the proposed novel robust loss can be easily adopted as a plug-in module for other methods.
	
	\section{Experiment}
	
	\subsection{Datasets}
	We conduct experiments on three long-tailed datasets: CIFAR-100-LT, ImageNet-LT, and iNaturalist 2018\footnote{\url{https://github.com/visipedia/inat_comp}}.
	
	CIFAR-100 is a balanced dataset containing 60,000 images from 100 categories. In our experiment, an imbalance factor $\lambda$, which is the ratio of sample numbers of the most frequent and least frequent classes, is used to generate different training sets for CIFAR-100-LT \cite{cui2019class,cao2019learning}. By varying $\lambda \in \left\lbrace 200,100,50,20,10\right\rbrace $, we can obtain five training sets. 
	
	ImageNet-LT is constructed from ImageNet by discarding some training samples \cite{liu2019large}. It has 115,846 training images, 20,000 validation images, and 50,000 testing images. The most frequent class contains 1,280 images, while the least frequent one only has 5 samples. 
	
	The iNaturalist 2018 is a large-scale species dataset with an extremely imbalanced distribution, where the imbalance factor is 500. It contains 437,513 training images and 24,426 validation images from 8,142 classes.

	\subsection{Results on CIFAR-100-LT}
	
	For CIFAR-100-LT, we use ResNet-32 as our backbone, which is trained by standard stochastic gradient descent (SGD) with a momentum of 0.9 and a weight decay $5\times10^{-4}$. We train the model for 200 epochs with a batch size of 100. The initial learning rate is set to 0.10, and the linear warm-up learning rate schedule is adopted.  Besides, we decay the learning rate by 0.01 at the $160^{th}$ and $180^{th}$ epochs. For the hyperparameters $\alpha$ and $\beta$, we select them from $\left\lbrace 0.25, 0.50, 0.75, 1.00,1.25,1.50\right\rbrace$. Different $\lambda$ have different optimal $\alpha$ and $\beta$, which are shown in Table \ref{tab0}. We compare our RISDA with some state-of-the-art methods. The results are shown in Table \ref{tab1}, where ``$^{\ast}$" indicates the results reported in the original paper, and ``$\dagger$" indicates the results reported in \cite{li2021metasaug}. 
	
	\begin{table}[!hbp]
		\centering
		\begin{tabular}{lccccc}
			\toprule
			$\lambda$    & 200 & 100 & 50 & 20 & 10 \\
			\midrule
			$\alpha$& 0.50    &  0.50   & 0.75   &  0.75  & 0.50   \\ 
			$\beta$&   1.00 &   0.75   & 1.00   & 0.75   & 0.50   \\ 
			\bottomrule
		\end{tabular}
		\caption{Optimal $\alpha$ and $\beta$ under different $\lambda$}
		\label{tab0}
	\end{table}


	
	We can draw the following conclusions: Firstly, our method can achieve the best performance compared with existing state-of-the-art methods. Secondly, the more unbalanced the data set is, the more our method improves. For example, when the $\lambda$ is 200 and 100, the error rate is reduced by 4.82$\%$ and 3.29 $\%$ compared with MetaSAug. While, when the $\lambda$ is 20, the error rate is reduced by 0.82$\%$. However, when the $\lambda$ is 10, the error rate is only reduced by 0.65$\%$. 
	We believe that this situation occurs because the fewer samples, the worse the covariance matrix learned by the tail category. Then, the variations transferred from similar categories can better improve the feature diversity of a tail category. Therefore, we can conclude that our method is more suitable for dealing with extremely imbalanced data.
	
	\begin{table}[t]
		\centering
		\begin{tabular}{l c}
			\toprule
			Method      & Error rate ($\%)$     \\ 
			\midrule
			Cross-Entropy $^{\dagger}$& 61.12                \\ 
			Class-balanced CE$^{\dagger}$ \cite{cui2019class}&59.15                      \\ 
			OLTR$^{\dagger}$ \cite{liu2019large}& 59.64                     \\ 
			LDAM$^{\dagger}$ \cite{cao2019learning}&58.14                      \\ 
			LDAM-DRW$^{\dagger}$ \cite{cao2019learning}&54.26                      \\ 
			Meta-class-weight$^{\dagger}$ \cite{jamal2020rethinking}&55.08                      \\ 
			MetaSAug$^{\ast}$ \cite{li2021metasaug} &52.61                      \\ 
			Bag of Tricks$^{\ast}$ \cite{zhang2021bag} &56.87                      \\ 
			\textbf{RISDA}&\textbf{49.31}                    \\ 
			\bottomrule
		\end{tabular}
		\caption{Results on ImageNet-LT of different methods. }
		\label{tab2}
	\end{table}

	\subsection{Results on ImageNet-LT }
	
	For fairness, we use ResNet-50 as the backbone. Concretely, the ResNet-50 is trained for 100 epochs by SGD with a momentum of 0.9 and a weight decay $2\times10^{-4}$. We set the initial learning rate to 0.1 and use the linear warm-up learning rate schedule for the first five epochs. Then, we decay the learning rate by 0.1 at the $60^{th}$ and $80^{th}$ epochs.
	Due to GPU memory limitation, we only transfer the covariance, so we augment a sample $\bm{f}_i$ during training with $N\left( \bm{f}_{i} ,\beta \left(\bm{\Sigma}_c+\bm{\Sigma}^{r}_{c}\right) \right)$. We set $\beta$ = 7.5. To recognize tail classes for augmentation, we define classes with more than 100 images as head classes, otherwise are the tail classes. 

	We compare our RISDA with the following methods: 1)Cross-Entropy(CE), class-balanced loss, meta-class-weight, LDAM-DRW, OLTR, and Bag of Tricks. 2)Mixup and MetaSAug. The experimental results are shown in Table \ref{tab2}. We can see that the error rate of RISDA is reduced to 49.31$\%$. More importantly, the error rate is reduced by 3.3$\%$ compared with MetaSAug, which shows that the diversity depicted by tail classes is limited, and can be greatly enriched by borrowing transformations from other classes.

	
	\subsection{Results on iNaturalist 2018}
	For iNaturalist 2018, we also implement ResNet-50 to achieve better classification results. We train the network with a batch size 128 for 120 epochs by SGD with a momentum of 0.9
	and a weight decay $1\times10^{-4}$. The learning rate is initialized to 0.05. We adopt the linear warm-up learning rate schedule \cite{goyal2017accurate} and decay the learning rate by 0.1 at $60^{th}$ and $80^{th}$ epochs. Referring to the setting in paper \cite{liu2019large}, we split the classes into head classes (with more than 100 images) and tail classes (with less than or equal to 100 images). Due to GPU memory limitation, we also only transfer the covariance, so a sample $\bm{f}_i$ is augmented during training with $N\left( \bm{f}_{i} ,\beta \left(\bm{\Sigma}_c+\bm{\Sigma}^{r}_{c}\right) \right)$.  
	\begin{table}
		\centering
		\begin{tabular}{l c}
			\toprule
			Method             & Error rate ($\%$)             \\ 
			\midrule
			Cross-Entropy$^{\dagger}$      &34.24            \\ 
			Class-balanced CE$^{\dagger}$ \cite{cui2019class}                    &33.57                      \\ 
			Class-balanced focal$^{\ast}$ \cite{cui2019class}                      &38.88                      \\ 
			cRT$^{\ast}$ \cite{kang2019decoupling}                      &32.40                      \\ 
			LDAM$^{\ast}$ \cite{cao2019learning}             & 35.42       \\ 
			LDAM-DRW$^{\ast}$ \cite{cao2019learning}   &32.00      \\ 
			BBN$^{\ast}$ \cite{zhou2020bbn}     & 33.71                      \\ 
			Meta-class-weight$^{\ast}$ \cite{jamal2020rethinking}                     &32.45                      \\ 
			MetaSAug$^{\ast}$ \cite{li2021metasaug}        &31.25      \\     
			Hybrid-PSC$^{\ast}$ \cite{wang2021contrastive}                     &31.90                      \\                
			Bag of Tricks$^{\ast}$ \cite{zhang2021bag}                    &\textbf{29.13}                     \\ 
			\textbf{RISDA}   & \multicolumn{1}{c}{   $\underline{30.85}$ } \\ 
			\bottomrule
		\end{tabular}
		\caption{Results on iNaturalist 2018 of different methods.}
		\label{tab3}
	\end{table}
	%
	
	We compare our RISDA with the following methods: 1) CE, class-balanced with CE and focal loss, meta-class-weight, and LDAM-DRW. 2) BBN, decoupling, and Hybrid-PSC. 3)Bag of Tricks and MetaSAug. The experimental results are shown in Table  \ref{tab3}. The average error rate of our method is reduced to 30.85$\%$.

	\subsection{Analysis}
	In this section, we conduct experiments to study the effect of different components of the proposed RISDA and discuss the parameter sensitivity on CIFAR-100-LT.
	
	\subsubsection{Ablation Study}
	
	To verify the effect of each part of our method, we do experiments by removing the re-weighting (w/o $w$)  or the reasoning (w/o $r$) component from our RISDA. Results shown in Table \ref{tab4} demonstrate that: (1) Re-weighting is important in our RISDA. Although we aim to generate unlimited samples through the Gaussian distribution, augmentation is implemented in each sample. Therefore, re-weighting plays a significant role to achieve fairness in training. (2) The reasoning module can effectively refine the feature and the covariance matrix to improve the accuracy of classification. (3) In addition, when $\lambda$ is 200, the reasoning module decreased by 1.73$\%$, and when the $\lambda$ is 10, the reasoning module decreased by 0.81$\%$, which proves again that our reasoning module is more suitable for dealing with extremely unbalanced data.
	
	
	\begin{table}[!ht]
		\centering
		\begin{tabular}{cccccc}
			\toprule
			$\lambda$ & 200   & 100   & 50    & 20    & 10    \\ 
			\midrule
			w/o $w$ & 36.82      &42.76       &46.98       &54.40       &58.60       \\ 
			w/o $r$    & 56.97 & 51.09 & 47.60  & 42.60  & 38.43 \\ 
			\textbf{RISDA}                                 &\textbf{55.24}       &\textbf{49.84}       &\textbf{46.16}       &\textbf{41.33}         &\textbf{37.62}       \\ 
			\bottomrule
		\end{tabular}
		\caption{Results of RISDA on CIFAR-100-LT under different imbalance factors.}
		\label{tab4}
	\end{table}
	
	
	\begin{figure*}[!tp]
		\centering
		\includegraphics[width=\linewidth]{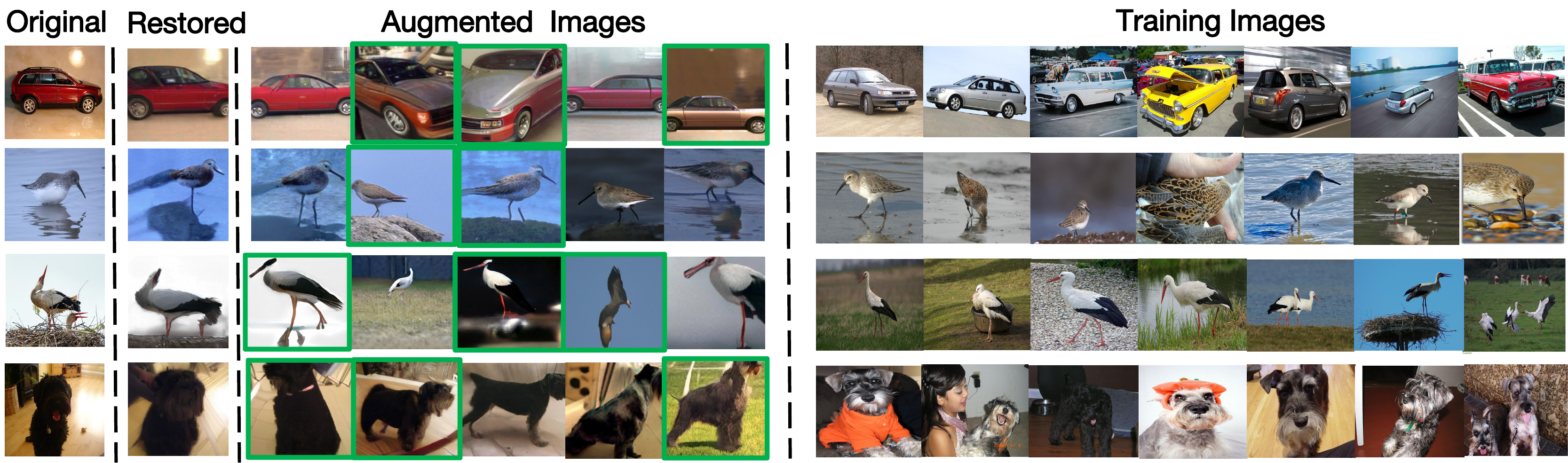}
		\caption{ Visualization of the semantically augmented images for the four tail classes on ImageNet-LT: Wagon, Dunlin, White Stork, and Miniature Schnauzer. ``Original" represents the original training sample. ``Restored" is the image reconstructed by a generator with features of the corresponding training sample. And ``Augmented Images" are generated with features sampled from our refined feature distribution. Our method can generate samples with semantic transformations that are not found in the original training set in terms of color, angle, pose, background, etc. These samples are highlighted in green boxes.}
		\label{visualization}
	\end{figure*}
	
	\subsubsection{Sensitivity of $\alpha$ and $\beta$}
	For the Gaussian distribution $N \left(\alpha \bm{\mu}_{c} ,\beta \left(\bm{\Sigma}_c+\bm{\Sigma}^{r}_{c}\right) \right)$ constructed by the reasoning module, the multivariate of sampling includes two significant componets: reasoning-based complementary feature $\alpha \bm{\mu}_{c}$ and the reasoning-based covariance matrix $\beta \left(\bm{\Sigma}_c+\bm{\Sigma}^{r}_{c}\right)$. We analyze the influence of the transformation strength $\alpha$, $\beta \in\left\lbrace 0.25,0.50,0.75,1.00,1.25,1.50\right\rbrace $. 
	\begin{figure}[!ht]
		\centering
		\includegraphics[width=0.95\linewidth]{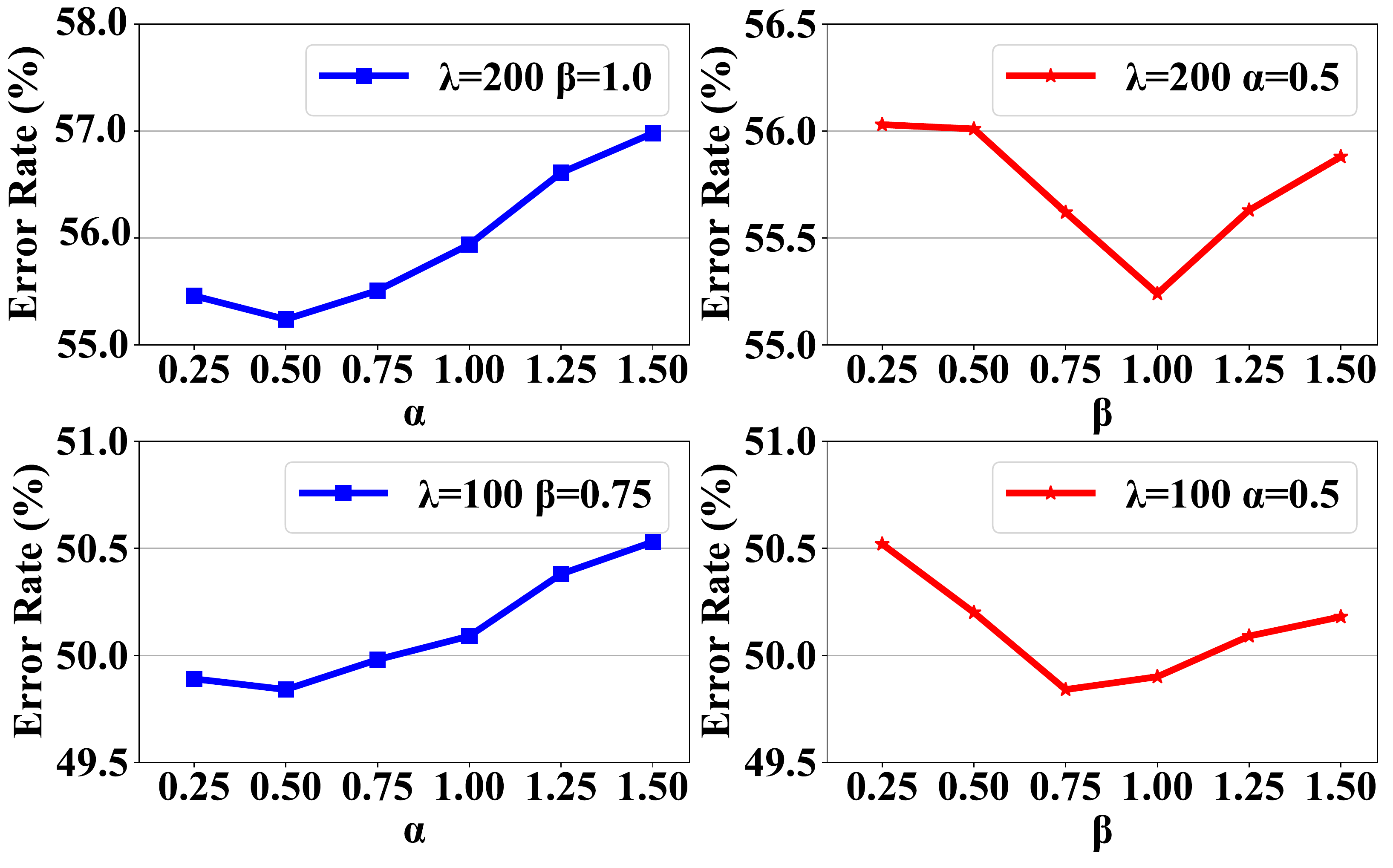}
		\caption{The influence of different $\alpha$ and $\beta$ on CIFAR-100-LT with $\lambda=$ 100 and $\lambda=$ 200.}
		\label{cxh_plot}
	\end{figure}

	As shown in Figure \ref{cxh_plot}, we can see that: (1) $\alpha$ = 0.50 can achieve the best performance with $\lambda$ = 100, and 200. Then, when $\alpha$ increases, the performance drops. It indicates that features of other classes can help to complement the feature of a tail instance. (2) $\beta$ = 1.00 achieves best results on $\lambda$ = 200, and $\beta$ = 0.75 performs best on $\lambda$ = 100. We can conclude that the more unbalanced, the larger $\beta$ is better.
	%
	\subsubsection{Effect of Head Categories}
	\begin{figure}
		\centering
		\includegraphics[width=0.95\linewidth]{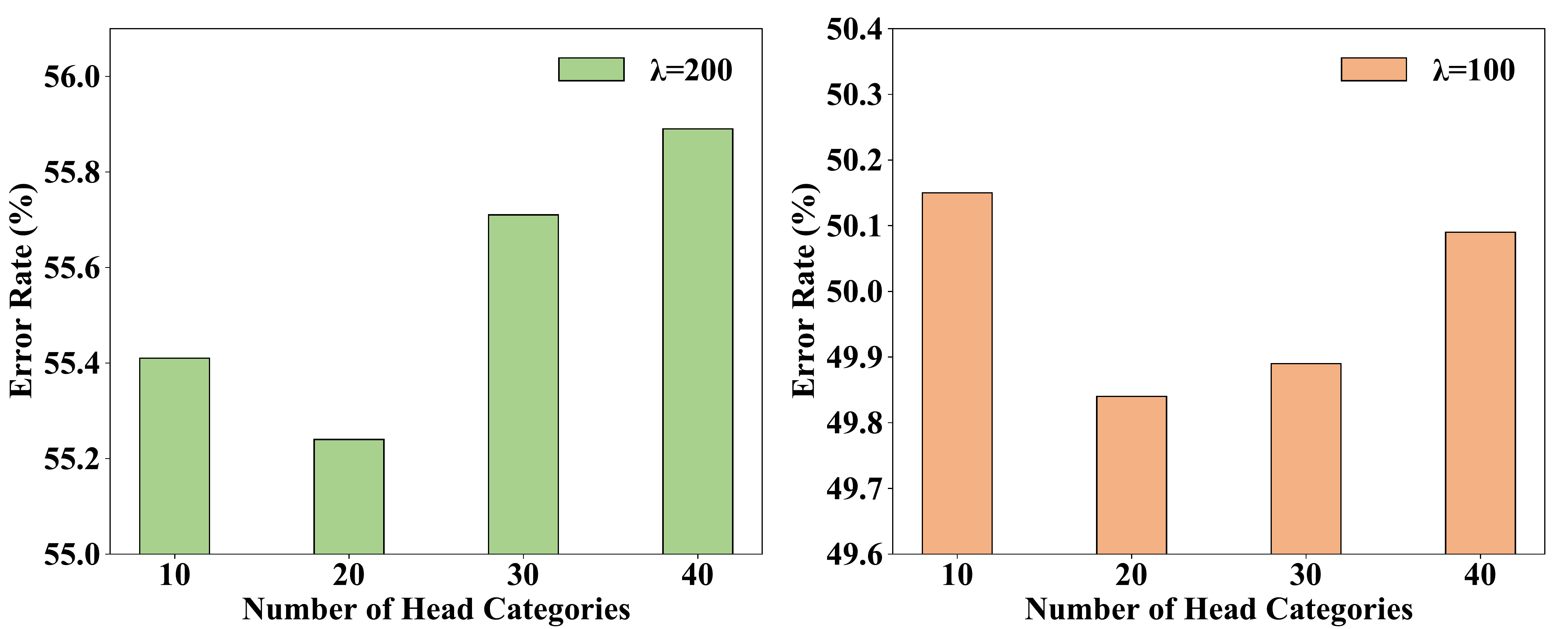}
		\caption{Results with different number of head categories.}
		\label{cxh_bar}
	\end{figure}
	Head categories are expected to have rich diversity. Transferring transformations from other classes may damage their representation, reducing the overall performance. Therefore, we explore the influence of head categories by varying the number in $\{10,20,30,40\}$.
	
	As shown in Figure \ref{cxh_bar}, the green bar on the left and the orange bar on the right shows the results with the different number of head categories on $\lambda = 200$ and $\lambda = 100$, respectively. We can see that when the top $20$ categories are regarded as head categories, the best performance is achieved. When it increases,  categories with limited diversity are underrepresented. However, when it decreases, the representation of some rich categories is damaged.
	

	\subsection{Qualitative Analysis}

	To intuitively show that our RISDA can generate more various samples, we implement the visualization method proposed in ISDA to show some implicitly generated samples. As shown in Figure \ref{visualization}, we can see that our RISDA can generate samples with different backgrounds, view angles, and poses. More importantly, images in green boxes are quite different from those in training sets. For example, the ``Wagon" changes the background, view angle, and color while the ``Dunlin" changed the posture, background, view angle.
	These changes are not available in the training data, and we believe they are delivered from similar categories. Therefore, it can be concluded that our RISDA can transfer more meaningful transformations than ISDA.
	
	\section{Conclusion and Future Work}
	In this paper, we have proposed a novel reasoning-based implicit semantic data augmentation method to effectively enrich the intra-class diversity for tail categories. By transferring transformation directions from other classes, an instance from tail categories can be augmented with definitely different semantics. Experimental results have shown that our proposed method can improve the performance of long-tail classification. Besides, the visualization experiment has demonstrated that our augmentation method can generate samples that are quite different from those in the training data. In the future, we will work on some fine-grained transformations to achieve more reasonable augmentation.

        \bibliography{aaai22}

\begin{thebibliography}{30}
\providecommand{\natexlab}[1]{#1}

\bibitem[{Bowles et~al.(2018)Bowles, Chen, Guerrero, Bentley, Gunn, Hammers,
  Dickie, del C.~Vald{\'{e}}s~Hern{\'{a}}ndez, Wardlaw, and
  Rueckert}]{bowles2018gan}
Bowles, C.; Chen, L.; Guerrero, R.; Bentley, P.; Gunn, R.~N.; Hammers, A.;
  Dickie, D.~A.; del C.~Vald{\'{e}}s~Hern{\'{a}}ndez, M.; Wardlaw, J.~M.; and
  Rueckert, D. 2018.
\newblock {GAN} Augmentation: Augmenting Training Data using Generative
  Adversarial Networks.
\newblock \emph{CoRR}, abs/1810.10863.

\bibitem[{Buda, Maki, and Mazurowski(2018)}]{buda2018systematic}
Buda, M.; Maki, A.; and Mazurowski, M.~A. 2018.
\newblock A systematic study of the class imbalance problem in convolutional
  neural networks.
\newblock \emph{Neural Networks}, 106: 249--259.

\bibitem[{Byrd and Lipton(2019)}]{byrd2019effect}
Byrd, J.; and Lipton, Z.~C. 2019.
\newblock What is the Effect of Importance Weighting in Deep Learning?
\newblock In \emph{{ICML}}, 872--881.

\bibitem[{Cao et~al.(2019)Cao, Wei, Gaidon, Ar{\'{e}}chiga, and
  Ma}]{cao2019learning}
Cao, K.; Wei, C.; Gaidon, A.; Ar{\'{e}}chiga, N.; and Ma, T. 2019.
\newblock Learning Imbalanced Datasets with Label-Distribution-Aware Margin
  Loss.
\newblock In \emph{NeurIPS}, 1565--1576.

\bibitem[{Chu et~al.(2020)Chu, Bian, Liu, and Ling}]{chu2020feature}
Chu, P.; Bian, X.; Liu, S.; and Ling, H. 2020.
\newblock Feature space augmentation for long-tailed data.
\newblock In \emph{ECCV}, 694--710.

\bibitem[{Cui et~al.(2019)Cui, Jia, Lin, Song, and Belongie}]{cui2019class}
Cui, Y.; Jia, M.; Lin, T.; Song, Y.; and Belongie, S.~J. 2019.
\newblock Class-Balanced Loss Based on Effective Number of Samples.
\newblock In \emph{{CVPR}}, 9268--9277.

\bibitem[{Cui et~al.(2018)Cui, Song, Sun, Howard, and
  Belongie}]{DBLP:conf/cvpr/CuiSSHB18}
Cui, Y.; Song, Y.; Sun, C.; Howard, A.; and Belongie, S.~J. 2018.
\newblock Large Scale Fine-Grained Categorization and Domain-Specific Transfer
  Learning.
\newblock In \emph{{CVPR}}, 4109--4118.

\bibitem[{Goyal et~al.(2017)Goyal, Doll{\'{a}}r, Girshick, Noordhuis,
  Wesolowski, Kyrola, Tulloch, Jia, and He}]{goyal2017accurate}
Goyal, P.; Doll{\'{a}}r, P.; Girshick, R.~B.; Noordhuis, P.; Wesolowski, L.;
  Kyrola, A.; Tulloch, A.; Jia, Y.; and He, K. 2017.
\newblock Accurate, Large Minibatch {SGD:} Training ImageNet in 1 Hour.
\newblock \emph{CoRR}, abs/1706.02677.

\bibitem[{He et~al.(2008)He, Bai, Garcia, and Li}]{he2008adaptive}
He, H.; Bai, Y.; Garcia, E.~A.; and Li, S. 2008.
\newblock {ADASYN:} Adaptive synthetic sampling approach for imbalanced
  learning.
\newblock In \emph{{WCCI}}, 1322--1328.

\bibitem[{He and Garcia(2009)}]{he2009learning}
He, H.; and Garcia, E.~A. 2009.
\newblock Learning from Imbalanced Data.
\newblock \emph{{IEEE} Trans. Knowl. Data Eng.}, 21(9): 1263--1284.

\bibitem[{He et~al.(2016)He, Zhang, Ren, and Sun}]{he2016deep}
He, K.; Zhang, X.; Ren, S.; and Sun, J. 2016.
\newblock Deep Residual Learning for Image Recognition.
\newblock In \emph{{CVPR}}, 770--778.

\bibitem[{Huang et~al.(2016)Huang, Li, Loy, and Tang}]{huang2016learning}
Huang, C.; Li, Y.; Loy, C.~C.; and Tang, X. 2016.
\newblock Learning deep representation for imbalanced classification.
\newblock In \emph{CVPR}, 5375--5384.

\bibitem[{Huang et~al.(2017)Huang, Liu, van~der Maaten, and
  Weinberger}]{HuangLMW17}
Huang, G.; Liu, Z.; van~der Maaten, L.; and Weinberger, K.~Q. 2017.
\newblock Densely Connected Convolutional Networks.
\newblock In \emph{{CVPR}}, 2261--2269.

\bibitem[{Jamal et~al.(2020)Jamal, Brown, Yang, Wang, and
  Gong}]{jamal2020rethinking}
Jamal, M.~A.; Brown, M.; Yang, M.; Wang, L.; and Gong, B. 2020.
\newblock Rethinking Class-Balanced Methods for Long-Tailed Visual Recognition
  From a Domain Adaptation Perspective.
\newblock In \emph{{CVPR}}, 7607--7616.

\bibitem[{Kang et~al.(2020)Kang, Xie, Rohrbach, Yan, Gordo, Feng, and
  Kalantidis}]{kang2019decoupling}
Kang, B.; Xie, S.; Rohrbach, M.; Yan, Z.; Gordo, A.; Feng, J.; and Kalantidis,
  Y. 2020.
\newblock Decoupling Representation and Classifier for Long-Tailed Recognition.
\newblock In \emph{{ICLR}}.

\bibitem[{Krizhevsky, Sutskever, and Hinton(2012)}]{KrizhevskySH12}
Krizhevsky, A.; Sutskever, I.; and Hinton, G.~E. 2012.
\newblock ImageNet Classification with Deep Convolutional Neural Networks.
\newblock In Bartlett, P.~L.; Pereira, F. C.~N.; Burges, C. J.~C.; Bottou, L.;
  and Weinberger, K.~Q., eds., \emph{Advances in Neural Information Processing
  Systems 25: 26th Annual Conference on Neural Information Processing Systems
  2012. Proceedings of a meeting held December 3-6, 2012, Lake Tahoe, Nevada,
  United States}, 1106--1114.

\bibitem[{Li et~al.(2021)Li, Gong, Liu, Wang, Qiao, and Cheng}]{li2021metasaug}
Li, S.; Gong, K.; Liu, C.~H.; Wang, Y.; Qiao, F.; and Cheng, X. 2021.
\newblock MetaSAug: Meta Semantic Augmentation for Long-Tailed Visual
  Recognition.
\newblock In \emph{{CVPR}}, 5212--5221.

\bibitem[{Lin et~al.(2017)Lin, Goyal, Girshick, He, and
  Doll{\'{a}}r}]{lin2017focal}
Lin, T.; Goyal, P.; Girshick, R.~B.; He, K.; and Doll{\'{a}}r, P. 2017.
\newblock Focal Loss for Dense Object Detection.
\newblock In \emph{{ICCV}}, 2999--3007.

\bibitem[{Liu et~al.(2020)Liu, Sun, Han, Dou, and Li}]{liu2020deep}
Liu, J.; Sun, Y.; Han, C.; Dou, Z.; and Li, W. 2020.
\newblock Deep representation learning on long-tailed data: A learnable
  embedding augmentation perspective.
\newblock In \emph{{CVPR}}, 2970--2979.

\bibitem[{Liu et~al.(2019)Liu, Miao, Zhan, Wang, Gong, and Yu}]{liu2019large}
Liu, Z.; Miao, Z.; Zhan, X.; Wang, J.; Gong, B.; and Yu, S.~X. 2019.
\newblock Large-scale long-tailed recognition in an open world.
\newblock In \emph{{CVPR}}, 2537--2546.

\bibitem[{Ratner et~al.(2017)Ratner, Ehrenberg, Hussain, Dunnmon, and
  R{\'{e}}}]{ratner2017learning}
Ratner, A.~J.; Ehrenberg, H.~R.; Hussain, Z.; Dunnmon, J.; and R{\'{e}}, C.
  2017.
\newblock Learning to Compose Domain-Specific Transformations for Data
  Augmentation.
\newblock In \emph{NeurIPS}, 3236--3246.

\bibitem[{Ren et~al.(2018)Ren, Zeng, Yang, and Urtasun}]{ren2018learning}
Ren, M.; Zeng, W.; Yang, B.; and Urtasun, R. 2018.
\newblock Learning to reweight examples for robust deep learning.
\newblock In \emph{ICML}, 4334--4343.

\bibitem[{Shu et~al.(2019)Shu, Xie, Yi, Zhao, Zhou, Xu, and Meng}]{shu2019meta}
Shu, J.; Xie, Q.; Yi, L.; Zhao, Q.; Zhou, S.; Xu, Z.; and Meng, D. 2019.
\newblock Meta-Weight-Net: Learning an Explicit Mapping For Sample Weighting.
\newblock In \emph{NeurIPS}, 1917--1928.

\bibitem[{Simonyan and Zisserman(2015)}]{simonyan2014very}
Simonyan, K.; and Zisserman, A. 2015.
\newblock Very Deep Convolutional Networks for Large-Scale Image Recognition.
\newblock In \emph{{ICLR}}.

\bibitem[{Wang et~al.(2021)Wang, Han, Wei, Zhang, and
  Wang}]{wang2021contrastive}
Wang, P.; Han, K.; Wei, X.-S.; Zhang, L.; and Wang, L. 2021.
\newblock Contrastive Learning based Hybrid Networks for Long-Tailed Image
  Classification.
\newblock In \emph{{CVPR}}, 943--952.

\bibitem[{Wang et~al.(2019)Wang, Pan, Song, Zhang, Huang, and
  Wu}]{wang2019implicit}
Wang, Y.; Pan, X.; Song, S.; Zhang, H.; Huang, G.; and Wu, C. 2019.
\newblock Implicit Semantic Data Augmentation for Deep Networks.
\newblock In \emph{NeurIPS}, 12614--12623.

\bibitem[{Xiao et~al.(2021)Xiao, Zhang, Jing, Huang, and Song}]{xiao2021does}
Xiao, L.; Zhang, X.; Jing, L.; Huang, C.; and Song, M. 2021.
\newblock Does Head Label Help for Long-Tailed Multi-Label Text Classification.
\newblock In \emph{{AAAI}}, 14103--14111.

\bibitem[{Zhang et~al.(2018)Zhang, Ciss{\'{e}}, Dauphin, and
  Lopez{-}Paz}]{zhang2017mixup}
Zhang, H.; Ciss{\'{e}}, M.; Dauphin, Y.~N.; and Lopez{-}Paz, D. 2018.
\newblock mixup: Beyond Empirical Risk Minimization.
\newblock In \emph{{ICLR}}.

\bibitem[{Zhang et~al.(2021)Zhang, Wei, Zhou, and Wu}]{zhang2021bag}
Zhang, Y.; Wei, X.; Zhou, B.; and Wu, J. 2021.
\newblock Bag of Tricks for Long-Tailed Visual Recognition with Deep
  Convolutional Neural Networks.
\newblock In \emph{{AAAI}}, 3447--3455.

\bibitem[{Zhou et~al.(2020)Zhou, Cui, Wei, and Chen}]{zhou2020bbn}
Zhou, B.; Cui, Q.; Wei, X.-S.; and Chen, Z.-M. 2020.
\newblock Bbn: Bilateral-branch network with cumulative learning for
  long-tailed visual recognition.
\newblock In \emph{{CVPR}}, 9719--9728.

\end{thebibliography}
		\begin{table*}[!pt]
		\centering
		\begin{tabular}{{lccccc}}
			\toprule
			Imbalance factor ($\lambda$)      & 200   & 100   & 50    & 20    & 10 \\
			\midrule
			Cross-Entropy    & 34.13& 29.86 & 25.06  & 17.56  & 13.82  \\
			Mixup$^{\dagger}$ \cite{zhang2017mixup}     &-       &26.94       &22.18       &-      &12.90       \\ 
			L2RW$^{\dagger}$ \cite{ren2018learning}        &33.75       &27.77      &23.55       &18.65       &17.88       \\ 
			Class-balanced CE$^{\dagger}$ \cite{cui2019class}         &31.23       &27.32       &21.87       &15.44      &13.10        \\ 
			Class-balanced  fine-tuning$^{\dagger}$ \cite{DBLP:conf/cvpr/CuiSSHB18}         &33.76      &28.66       &22.56       &16.78       &16.83       \\ 
			Meta-weight net$^{\dagger}$ \cite{shu2019meta}    &32.80     &26.43      &20.90      &15.55       &12.45       \\ 
			Meta-class-weight with cross-entropy loss$^{\dagger}$ \cite{jamal2020rethinking} &29.34       &23.59     &19.49       &13.54       &11.15       \\ 
			BBN$^{\ast}$ \cite{zhou2020bbn}    &-       &20.18       &17.82       & -       &11.68       \\ 
			Hybrid-PSC$^{\ast}$ \cite{wang2021contrastive}                                &-       &21.18       & \textbf{14.64}      &-      &\textbf{9.94}       \\
			Bag of Tricks$^{\ast}$ \cite{zhang2021bag}                           &-       &19.97       &16.41       &-       &-       \\ 
			MetaSAug with cross-entropy loss$^{\ast}$ \cite{li2021metasaug}          &\textbf{23.11}    & \textbf{19.46}       &15.97       &\textbf{ 12.36}       &10.56       \\ 
			\textbf{RISDA}                                 &\multicolumn{1}{c}{   $\underline{26.00}$ }       &\multicolumn{1}{c}{   $\underline{20.11}$ }        &\multicolumn{1}{c}{   $\underline{15.76 }$ }         &\multicolumn{1}{c}{   $\underline{13.02}$ }          &\multicolumn{1}{c}{   $\underline{10.64}$ }      \\
			\bottomrule
		\end{tabular}
		\caption{Error rate ($\%$) of ResNet-32 on CIFAR-10-LT under different imbalance factors.}
		\label{cifar10}
	\end{table*}
\newpage


\section*{Appendix}

\subsection{Details about Baselines on CIFAR}

In the CIFAR-100-LT and CIFAR-10-LT experiment, we compare with the following methods: (1) Cross-Entropy (CE): we train the model with vanilla cross-entropy loss.
(2)Re-weighting (RW): we train the model with two re-weighting strategies: class-level and instance-level re-weighting. Class-level re-weighting assign different weights to different classes according to the number of samples in loss function, including class-balanced loss \cite{cui2019class} , meta-class-weight \cite{jamal2020rethinking}, and LDAM-DRW \cite{cao2019learning}. Instance-level re-weighting allocate weights to samples, including Focal loss \cite{lin2017focal},  L2RW \cite{ren2018learning} and meta-weight-net \cite{shu2019meta}. (3)Data augmentation method improves classification performance by enhancing data diversity, including mixup \cite{zhang2017mixup} and MetaSAug \cite{li2021metasaug}. 
(4)Two-stage learning splits the learning procedure into feature representation learning and classifier training, such as BBN \cite{zhou2020bbn}, decoupling \cite{kang2019decoupling} and Hybrid-PSC \cite{wang2021contrastive}. In addition, we also compare with Bag of Tricks\cite{zhang2021bag}, which integrates the commonly used training tricks in long-tail classification.

\subsection{Results on CIFAR-10-LT}

CIFAR-10 is a balanced dataset containing 60,000 images from 10 categories, and the training set has 5000  images per class. In our experiment, an imbalance factor $\lambda$, which is the ratio of sample numbers of the most frequent and least frequent classes, is used to generate different training sets for CIFAR-10-LT \cite{cui2019class,cao2019learning}. By varying $\lambda \in \left\lbrace 200,100,50,20,10\right\rbrace $, we can obtain five training sets. A long-tail distribution usually contains many classes. As CIFAR-10-LT only contains 10 classes, we leave out its experiments in the main body. 

Experimental results on CIFAR-10-LT are shown in Table \ref{cifar10}. 
We can see that our RISDA is better than most baseline methods, but achieves comparable performance with MetaSAug. We believe that this situation occurs because our method is good at dealing with large-scale long-tailed classification problem. With more categories, there are rich similarity relations to make reasonable transformatnion. In addition, the CIFAR-10-LT dataset contains a large number of samples in each category. Even in the most unbalanced case, there are 25 samples for the least tail categories. Consequently, the diversity of samples learned in each category is rich enough. Therefore, although RISDA performs better, we argue that it is more suitable to deal with many classes.
\subsection{Comparison with ISDA}
Our RISDA is developed from ISDA to achieve implicitly data augmentation. However, ISDA fails on long-tail distributed data. Firstly, the diversity of augmented data for tail classes is quite limited because ISDA enhances a sample with the covariance of its class. As tail classes only own a few samples, their covariance is limited. Secondly, ISDA lacks strategies to achieve balance training. For the sake of fairness, we show the results of ISDA and ISDA with reweighting in Table \ref{cifar100}. We can see that even if equipped with reweighting, ISDA is still worse than RISDA because of the limited diversity.
\begin{table}[!ht]
	\centering
	\begin{tabular}{lclclclclclc}
		\toprule
		$\lambda$      & 200   & 100   & 50    & 20    & 10 \\
		\midrule
		CE$^{\dagger}$      & 65.30 & 61.54 & 55.98 & 48.94 & 44.27 \\ 
		ISDA  & 64.48      &58.65       &54.35      &47.25       &42.84      \\
		ISDA+$w$     & 56.97 & 51.09 & 47.6  & 42.6  & 38.43  \\
		\textbf{RISDA}                                 &\textbf{55.24}       &\textbf{49.84}       &\textbf{46.16}       &\textbf{41.33}         &\textbf{37.62}       \\
		\bottomrule
	\end{tabular}
	\caption{Error rates ($\%$) on CIFAR-100-LT.}
	\label{cifar100}
\end{table}

Our RISDA diversify tail classes from two aspects: (1) We transfer covariance from similar classes to tail classes to expand the variations. Therefore, RSIDA can imagine a flying bird when there are only birds on the roost in the training set(the flying is transferred from similar classes). (2) A tail sample may miss some features since tail classes are under-fitted. If a bird's wing features can't be extracted, the corresponding semantic changes will not work. In RSIDA, we can infer some missing features through the knowledge graph, so as to achieve more variations for a specific sample.

\end{document}